\documentclass[10pt,twocolumn,letterpaper]{article}

\usepackage{cvpr}      % To produce the REVIEW version
\definecolor{cvprblue}{rgb}{0.21,0.49,0.74}
\usepackage[pagebackref,breaklinks,colorlinks,allcolors=cvprblue]{hyperref}
\usepackage{algorithm}    
\usepackage{algorithmic}
\usepackage{multirow}
\usepackage{booktabs}
\usepackage{float}
\usepackage{graphicx}
\usepackage{pifont}
\usepackage{xcolor,colortbl}
\newcommand{\cmark}{\ding{51}}%
\def\paperID{*****} % *** Enter the Paper ID here
\def\confName{CVPR}
\def\confYear{2026}

\title{Beyond Prompt Degradation: Prototype-guided Dual-pool Prompting for Incremental Object Detection}

\author{
Yaoteng Zhang, Qing Zhou, Junyu Gao, Qi Wang\thanks{Corresponding author.}\\
Northwestern Polytechnical University\\
{\tt\small zhang\_yt@mail.nwpu.edu.cn,
mrazhou@mail.nwpu.edu.cn,
gjy3035@gmail.com,
crabwq@gmail.com}
}

\begin{document}
\maketitle
\begin{abstract}
Incremental Object Detection (IOD) aims to continuously learn new object categories without forgetting previously learned ones. Recently, prompt-based methods have gained popularity for their replay-free design and parameter efficiency. However, due to prompt coupling and prompt drift, these methods often suffer from prompt degradation during continual adaptation.
To address these issues, we propose a novel prompt-decoupled framework called PDP. PDP innovatively designs a dual-pool prompt decoupling paradigm, which consists of a shared pool used to capture task-general knowledge for forward transfer, and a private pool used to learn task-specific discriminative features.
This paradigm explicitly separates task-general and task-specific prompts, preventing interference between prompts and mitigating prompt coupling.
In addition, to counteract prompt drift resulting from inconsistent supervision where old foreground objects are treated as background in subsequent tasks, PDP introduces a Prototypical Pseudo-Label Generation (PPG) module. PPG can dynamically update the class prototype space during training and use the class prototypes to further filter valuable pseudo-labels, maintaining supervisory signal consistency throughout the incremental process.
PDP achieves state-of-the-art performance on MS-COCO (with a 9.2\% AP improvement) and PASCAL VOC (with a 3.3\% AP improvement) benchmarks, highlighting its potential in balancing stability and plasticity. The code
and dataset are released at: https://github.com/zyt95579/PDP\_IOD/tree/main

\end{abstract}    
\section{Introduction}
\label{sec:intro}

\begin{figure}[t]
  \centering
   \includegraphics[width=0.9\linewidth]{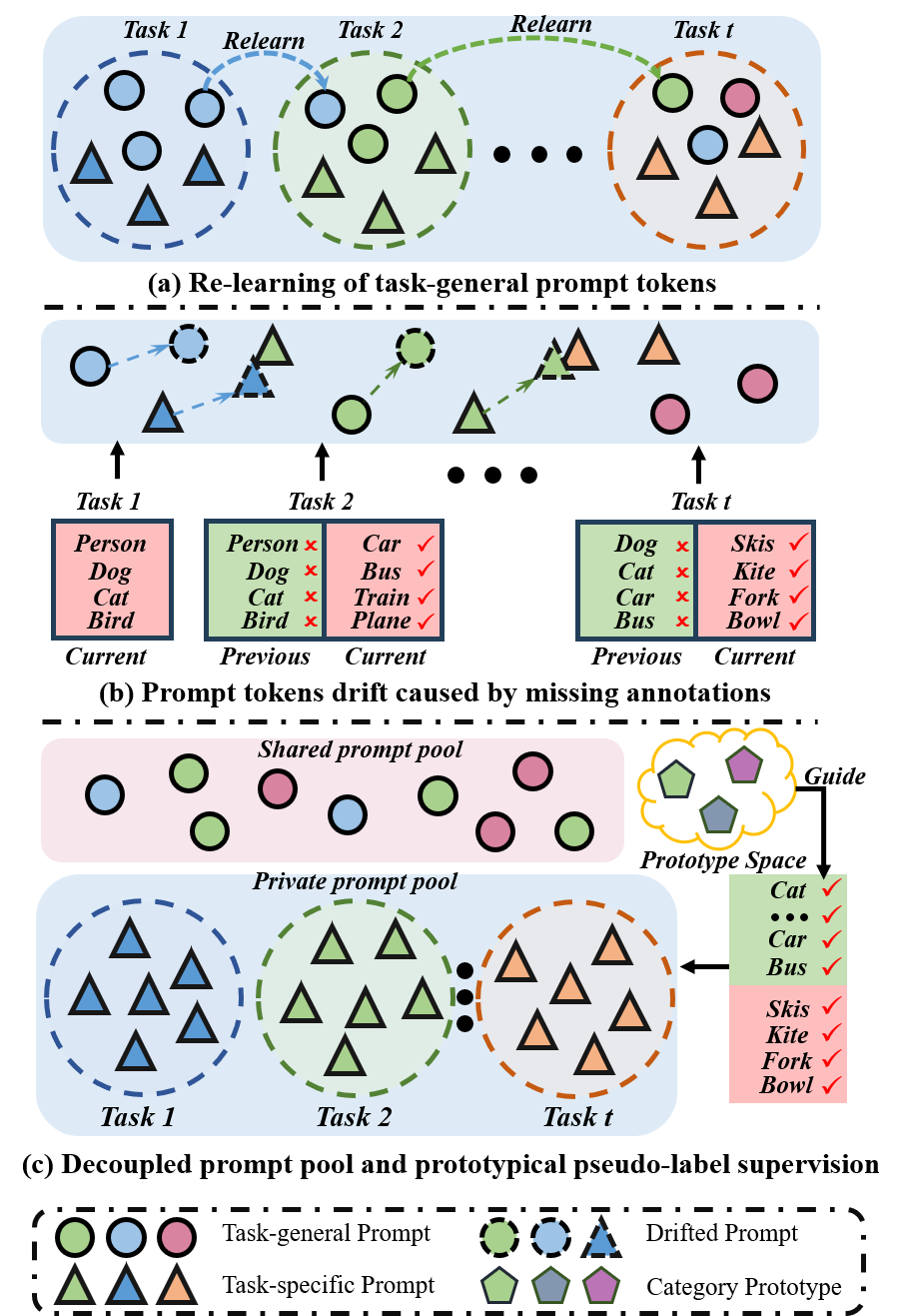}

   \caption{Comparison of prompt-based methods. (a) Use task ID to isolate prompt, which forces task-general prompt to be relearned at each task. (b) Missing annotations cause prompt tokens to drift. (c) The shared prompt pool continuously optimizes task-general prompt, and category prototypes guide the generation of pseudo labels for old categories.}
   \label{fig:onecol}
\end{figure}
\hspace*{1em}Incremental Object Detection (IOD) aims to continuously learn and recognize new object categories from a sequential data stream, while maintaining detection performance on previously learned classes without relying on past data~\cite{shmelkov2017incremental,mo2024bridge,li2017learning,Sun_2025_CVPR,Gao2024NWPUMOCAB,10530145,Zhou2025ScaleET}. This setting embodies the stability–plasticity dilemma: a model must remain plastic enough to integrate novel knowledge yet stable enough to avoid catastrophic forgetting. Continual learning (CL) aims to mitigate catastrophic forgetting by achieving a balance between stability and plasticity, thereby enabling continuous adaptation to open-world tasks~\cite{bhatt2024preventing,liu2023augmented,zhang2024not,Zhu_2025_CVPR,rim2025protodepthunsupervisedcontinualdepth,10902491,11083637,li2025exploring,liu2025msdp}.

Recently, prompt-based methods~\cite{hong2025rainbowprompt,roy2024convolutional,liu2025sec,jiang2025revisitingpoolbasedpromptlearning} have emerged as a promising paradigm for IOD, effectively avoiding data replay~\cite{rebuffi2017icarl,bang2021rainbow,aljundi2019gradient,isele2018selective,iscen2020memory,liu2020mnemonics} and uncontrolled model expansion~\cite{ostapenko2019learning,yoon2017lifelong,xu2018reinforced,li2019learn,yan2021dynamically,wang2022foster}. 
Despite their success, existing prompt-based continual learning methods still suffer from \textbf{prompt degradation}. We find that such degradation manifests as \textbf{prompt coupling} and \textbf{prompt drift}.

Specifically, current prompt-based methods~\cite{wang2022learning,smith2023coda,wang2022dualprompt,bhatt2024preventing} typically follow a single prompt pool paradigm.
\textbf{Prompt coupling} arises from this paradigm’s indiscriminate storage of functionally distinct prompts—shareable task-general prompts and discriminative task-specific prompts—within the same pool.
Such a design forces them to compete and interfere under limited parameter space, causing prompt degradation, as illustrated in Fig.~\ref{fig:onecol}(a).
Furthermore, the inconsistency in supervisory labels leads to \textbf{prompt drift}.
In IOD settings, when learning new tasks, previously learned objects are relabeled as “background”.
This inconsistent supervision forces the model to update prompts that have already been optimized for old tasks, causing them to drift toward incorrect semantic directions, as shown in Fig.~\ref{fig:onecol}(b).
Although some methods~\cite{sohn2020fixmatch,bang2021rainbow,zhang2024decoupled,yan2025ucod,zou2025mosaic} attempt to compensate for the inconsistent supervision through pseudo-labels, they generally rely on a static confidence threshold, which is unsuitable for continual learning. Due to the category-wise distribution discrepancy, such fixed thresholds become ill-calibrated, producing unstable supervision that may further exacerbate prompt drift.

To address these challenges, particularly \textbf{\textit{the representation degradation caused by prompt coupling and prompt drift}}, we propose PDP, a framework characterized by \textbf{\textit{prompt decoupling}} and \textbf{\textit{prototype-guided pseudo-labeling mechanism}}.
Specifically, to ensure that prompts remain decoupled throughout cross-task learning, PDP introduces a dual-pool paradigm consisting of a shared pool and a private pool, which explicitly separates task-general prompts and task-specific prompts, as illustrated in Fig.~\ref{fig:onecol}(c). The shared pool is continuously optimized to capture generalized visual knowledge, facilitating stable forward knowledge transfer, while the private pool preserves task-specific representations, effectively preventing knowledge forgetting.
This dual-pool design ensures that task-general and task-specific prompts evolve collaboratively yet independently during continual learning, thereby preventing prompt degradation.
Moreover, to counteract prompt drift induced by supervision inconsistency, where old foreground instances are incorrectly treated as background in later tasks, PDP introduces a Prototypical Pseudo-Label Generation (PPG) module.
During training, PPG constructs a category prototype space and generates pseudo-labels by measuring the distance between candidate samples (even those with low confidence) and the corresponding class prototypes in the embedding space, ensuring consistent supervision and preventing prompt drift throughout continual learning.

In summary, the main contributions of this work are as follows:

\begin{itemize}
\item To the best of our knowledge, we are the first to propose a dual prompt-pool framework for IOD that explicitly decouples task-general from task-specific prompts, significantly enhancing the model stability-plasticity balance.

\item We design a Prototypical Pseudo-label Generation (PPG) that employs prototype-to-feature similarity in the embedding space to produce reliable and semantically consistent pseudo labels.

\item We achieve SOTA performance across multiple MS-COCO and Pascal VOC incremental settings, with 59.4\% and 79.4\% mAP, respectively.
\end{itemize}
%-------------------------------------------------------------------------

\section{Related Work}
\label{sec:formatting}
\begin{figure*}[t]
  \centering
   \includegraphics[width=0.8\linewidth]{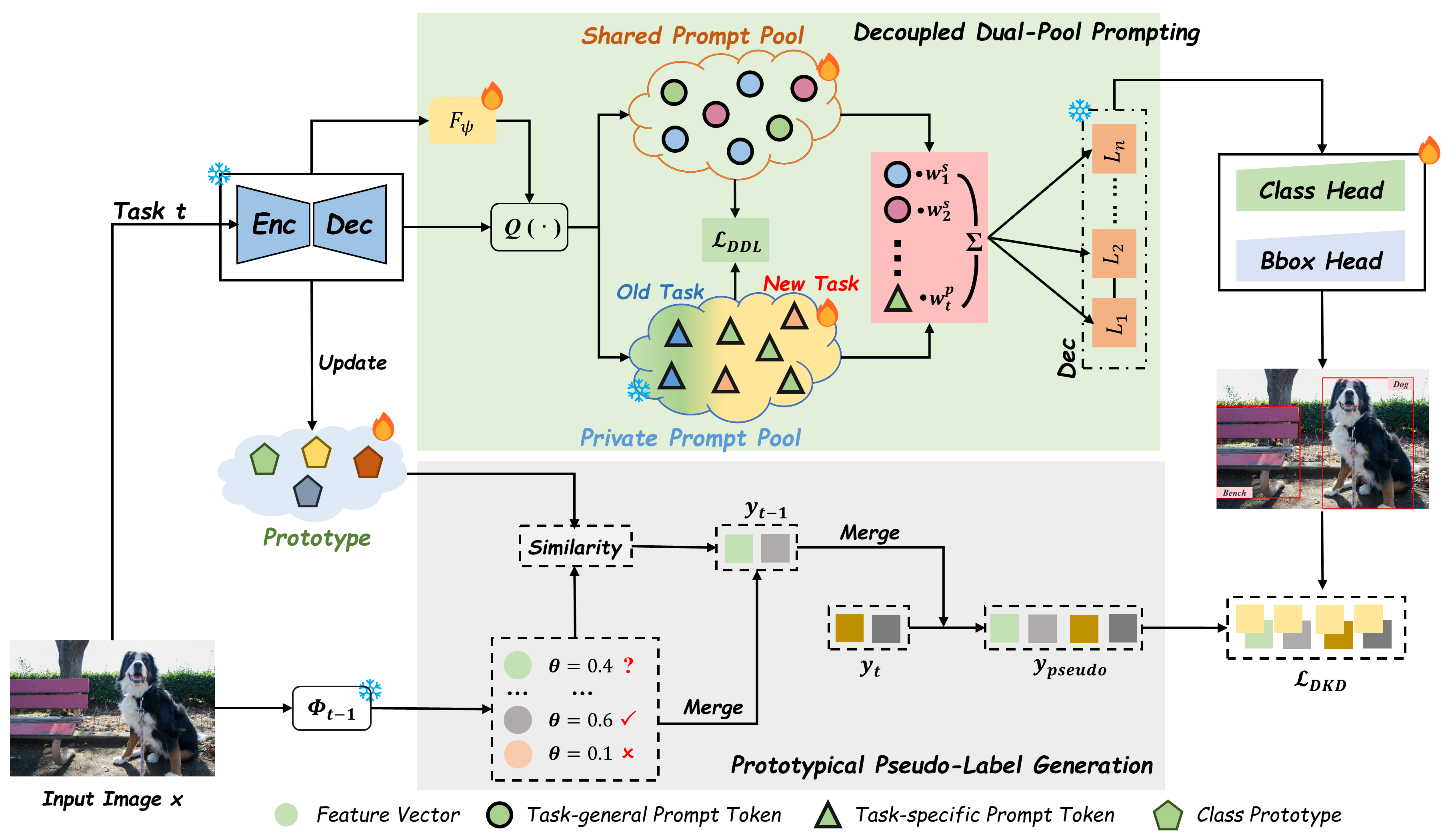}
   \caption{Overview of our framework at incremental step $t$. 
   Given an image $x$, the query function generates a content-aware query representation by adaptively computing query weights via a ranking function $F_\psi$ and performing weighted aggregation. Subsequently, prompts are retrieved from both the shared and private pool and injected into the decoder layer. In parallel, the teacher model $\Phi_{t-1}$ generates a set of candidate bounding boxes, where potentially valuable ones are projected into the feature space to compute their similarity with class prototypes. This process yields a set of refined, high-quality pseudo-labels to guide the training of the student model $\Phi_t$.}
   \label{fig:over}
\end{figure*}
\noindent\textbf{Prompt-based Continual Learning Methods.} 
Prompt-based approaches have recently become a prevailing paradigm in Continual Learning, providing an effective way to mitigate catastrophic forgetting without exemplar replay. L2P~\cite{wang2022learning} introduced a shared prompt pool, where the top-$K$ prompts are selected to preserve prior knowledge. In contrast, Coda-Prompt~\cite{smith2023coda} and MD-DETR~\cite{bhatt2024preventing} use task IDs to isolate prompt parameters. During training, only prompt parameters for new tasks are updated, while prompt parameters for old tasks remain frozen. The inference process retrieves prompts by calculating similarity. However, this rigid design constrains knowledge reuse and adaptation. DualPrompt~\cite{wang2022dualprompt} further partitions prompts into General-Prompts and Expert-Prompts for different decoder layers, yet both are still managed within a single pool—hindering proper decoupling and specialization of task knowledge. ConvPrompt~\cite{roy2024convolutional} generates prompts via convolution over a shared pool combined with external text descriptors, but its performance heavily depends on the quality of textual inputs. CPrompt~\cite{gao2024consistent} adopts a component-growth strategy by incrementally adding task-specific prompts, with consistency regularization to preserve past knowledge. Despite these advances, prompt-based methods still use a single prompt pool to jointly model task-specific and task-general prompts, leading to mutual interference among prompts and resulting in prompt degradation.

\noindent\textbf{Incremental Object Detection.} 
IOD particularly challenging since missing annotations of old classes cause them to be misclassified as background. ILOD~\cite{shmelkov2017incremental} employed the Learning without Forgetting (LwF)~\cite{li2017learning} strategy to mitigate forgetting. ABR~\cite{liu2023augmented} integrates old-class bounding box replay with RoI distillation, yielding improved retention of past object representations. In addition, some methods focus on generating pseudo-labels to prevent the forgetting of old knowledge. OW-DETR~\cite{gupta2022ow} introduced an attention-based pseudo-label generation mechanism to discover unannotated old-class instances. CL-DETR~\cite{liu2023continual} and MD-DETR~\cite{bhatt2024preventing} incorporated threshold-based pseudo-label supervision to alleviate forgetting. However, due to the significant differences in confidence distributions across different classes, a fixed threshold cannot be effectively generalized to all classes. To address this, PseDet~\cite{wangpsedet} proposed a categorical adaptive label selector that dynamically adjusts thresholds for each category via k-means clustering on confidence scores, achieving more accurate and balanced pseudo-label selection. However, PseDet essentially performs classification on confidence distributions and is not an end-to-end framework. Unlike confidence-based pseudo-labeling, we propose a prototype-guided pseudo-label generation module that utilizes category prototypes in the embedding space to guide pseudo-label generation.

%-------------------------------------------------------------------------
\section{Preliminaries}

\noindent\textbf{Incremental Object Detection.} 
The training process of IOD is organized into $n$ sequential stages, where each stage introduces a disjoint set of novel categories. Let the overall class set be $\mathcal{C} = \{C_1, C_2, \ldots, C_t, \ldots, C_n\}$ with $C_i \cap C_j = \emptyset, \forall i \neq j$. At stage $t$, the detector is trained on the dataset $\mathcal{D}_t = \{X_t, Y_t\}$, where $X_t$ represents the input images and $Y_t$ the corresponding annotations of classes in $C_t$. Although images may contain objects from any category in $\mathcal{C}$, only instances of $C_t$ are annotated. The objective is to adapt the detector from $\mathcal{M}_{t-1}$ to $\mathcal{M}_t$ using only the current dataset $\mathcal{D}_t$, without accessing previous datasets $\{\mathcal{D}_1, \ldots, \mathcal{D}_{t-1}\}$, while preventing catastrophic forgetting and maintaining strong performance on previously learned classes $\{C_1, \ldots, C_{t-1}\}$.

\noindent\textbf{MD-DETR.} 
MD-DETR~\cite{bhatt2024preventing} introduces a learnable memory bank that stores key–memory pairs for information retrieval. Its architecture consists of a frozen encoder–decoder $\Theta_{\nabla}$ and a localized query function $Q$ guided by a ranker $g_{\psi}$: 
\begin{equation}
Q(x, \Theta_{\nabla}, \alpha) = \sum_i \alpha_i \cdot \{\Theta_{\nabla}(x)\}_i
\end{equation}

The relevance weights $\alpha$ are computed by $g_{\psi}(\Theta_{\nabla}(x))$. The resulting query $Q(x, \Theta_{\nabla}, \alpha)$ then interacts with the keys in the memory bank to compute similarity scores, which are used to aggregate the corresponding memory values into a retrieved representation. This retrieved memory is then injected into the decoder through prefix-tuning~\cite{wang2022dualprompt}. The overall training loss combines the DETR detection loss and a retrieval regularization:
\begin{equation}
\mathcal{L}_{{MD-DETR}} = \mathcal{L}_{{DETR}} + \mathcal{L}_Q
\end{equation}

Given ground-truth pairs $Y = \{(c_i, b_i)\}_{i=1}^M$ and predictions $\hat{Y} = \{(\hat{s}_j, \hat{b}_j)\}_{j=1}^N$. The optimal one-to-one assignment $\hat{\sigma}$ is obtained by minimizing the matching cost:
\begin{equation}
\hat{\sigma} = \arg\min_{\sigma \in S_N} \sum_{i=1}^M \mathcal{L}_{{match}}\big((c_i, b_i), (\hat{s}_{\sigma(i)}, \hat{b}_{\sigma(i)})\big)
\end{equation}
and the detection loss is formulated as:
\begin{equation}
\mathcal{L}_{{DETR}} = \sum_i \Big[\mathcal{L}_{{cls}}(c_i, \hat{s}_{\hat{\sigma}(i)}) 
+ \mathbb{I}_{\{c_{\hat{\sigma}(i)} \neq \emptyset\}} \cdot \mathcal{L}_{{box}}(b_i, \hat{b}_{\hat{\sigma}(i)})\Big]
\end{equation}

$\mathcal{L}_Q$ encourages the ranker $g_{\psi}$ to produce relevance scores consistent with the optimal assignment $\hat{\alpha}$ obtained via Hungarian matching, implemented as a cross-entropy loss:
\begin{equation}
\mathcal{L}_Q = \lambda_Q \cdot \mathcal{L}_{{CE}}(\alpha, \hat{\alpha})
\end{equation}

\section{Method}

\subsection{Overall Framework}

As illustrated in Fig.~\ref{fig:over}, PDP is a fully end-to-end trainable framework that effectively prevents prompt degradation through prompt decoupling and pseudo-label generation. The overall framework of PDP consists of two core modules: Decoupled Dual-Pool Prompting (DDP) and Prototypical Pseudo-Label Generation (PPG).
Specifically, DDP decouples the prompts into task-general prompts and task-specific prompts, which are independently updated through a shared pool and a private pool, achieving explicit knowledge disentanglement and collaborative optimization.
PPG avoids reliance on confidence thresholds and instead guides pseudo-label generation through a dynamically updated prototype space.

\subsection{Decoupled Dual-Pool Prompting}
DDP maintains two distinct prompt pools — a shared pool and a private pool — to explicitly decouple task-general and task-specific prompt representations.
This design prevents cross-task interference, allowing the model to acquire discriminative task-specific knowledge while preserving reusable global information.
Furthermore, DDP enforces an inter-pool diversity constraint to ensure that shared and private prompts learn complementary, orthogonal representations.

\noindent\textbf{Shared Pool.} 
The shared pool serves as a global repository of transferable prompts accessible to all tasks.
It is composed of learnable prompt tokens $P_s \in \mathbb{R}^{N_s \times L_p \times D}$, key vectors $K_s \in \mathbb{R}^{N_s \times D}$, and a query adapter $A_s \in \mathbb{R}^{N_s \times D}$. 
This pool is designed to encode reusable visual prompts transferable across tasks.
During each incremental stage, it is progressively refined with data from new tasks, thereby enhancing its representational capacity.
As a globally accessible component, the shared pool facilitates stable forward knowledge transfer throughout the training process.

\begin{figure*}[t]
\centering
\includegraphics[width=0.8\linewidth]{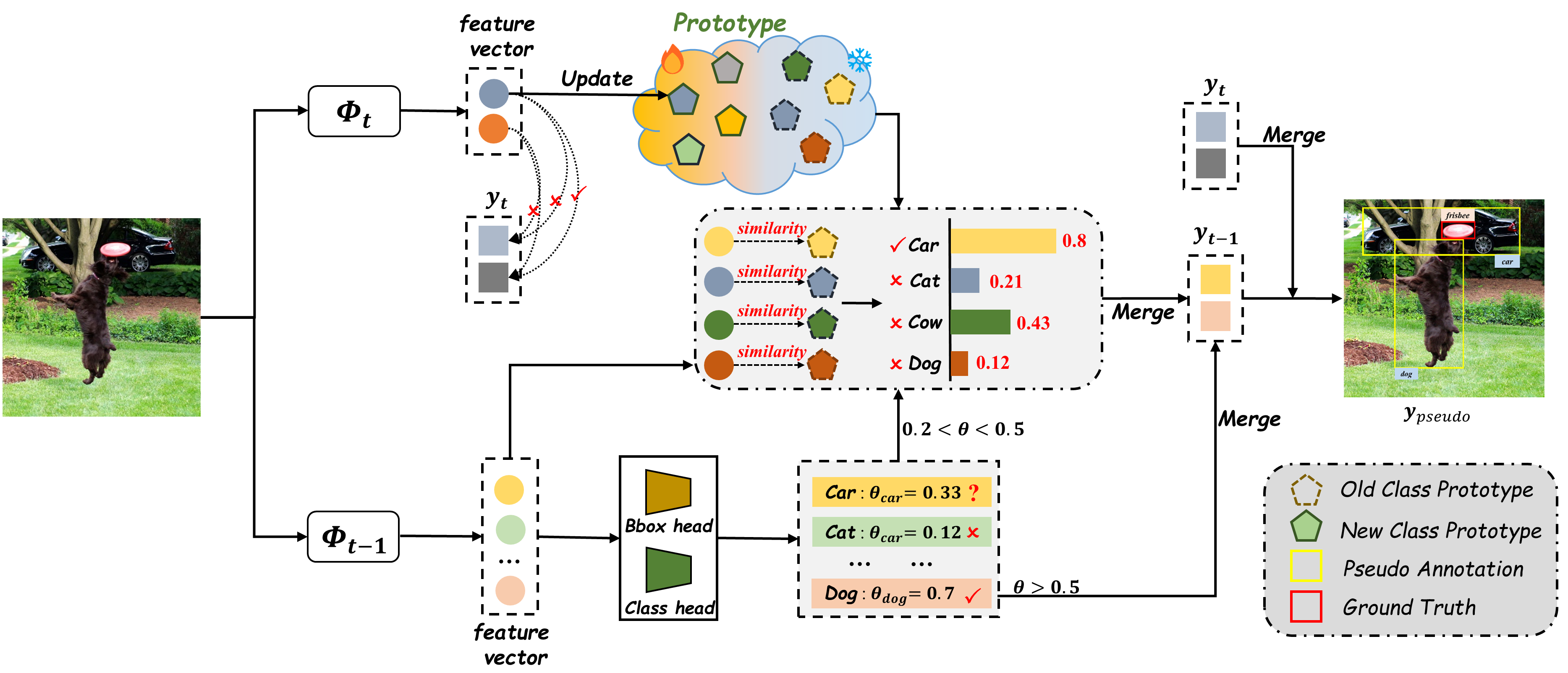}
\caption{
The pseudo-label generation process of PPG at stage $t$. 
PPG dynamically updates the prototypes of new task classes while keeping the old class prototypes frozen. 
The teacher model $\Phi_{t-1}$ produces a set of candidate detections, where high-confidence predictions are directly regarded as reliable samples. 
For low-confidence candidates, similarity matching with frozen old class prototypes is performed, and those exceeding a predefined threshold are also considered reliable. 
Finally, both types of samples are merged to generate high-quality pseudo-labels.
}
\label{fig:ppr}
\end{figure*}

\noindent\textbf{Private Pool.} 
Preserving task-specific discriminative information is crucial for preventing catastrophic forgetting. To this end, we introduce a private pool that dynamically maintains task-specific prompts.
For each task, a set of prompt parameters $(P_p, K_p, A_p)$ is privatized to retain task-specific representations. 
During training on task $t$, only the current task prompt parameters $(P_p^{t}, K_p^{t}, A_p^{t})$ are trainable, while the prompt parameters from previous tasks $\theta_{i<t}^{priv}$ are frozen. 
Unlike the shared pool, the number of prompt tokens $N_p$ in the private pool is dynamically adjusted based on the number of new classes introduced in each task.
The private pool not only prevents interference among tasks but also provides sufficient representational capacity for newly introduced classes.

\noindent\textbf{Prompt Retrieval and Integration.}
Given an input image $I$, the query extractor $Q(\cdot)$ generates a query vector \( Q(x, \theta_\nabla):\mathbb{R}^{3 \times H \times W} \rightarrow \mathbb{R}^{1 \times D}\). 
The retrieval process begins by modulating the query with the query adapter through an element-wise Hadamard product, followed by computing cosine similarity scores $\rho$ with both key vectors $(K_s, K_p)$:
\begin{equation}
w = \rho\big(Q(x, \theta_\nabla) \odot [A_s, A_p], [K_s, K_p]\big)
\end{equation}
\begin{equation}
P_r = \sum_{i=1}^{N_s} w_s^i P_s^i + \sum_{j=1}^{N_p} w_p^j P_p^j
\end{equation}
The retrieved prompt $P_r$ acts as a contextual memory token, aggregating task-general and task-specific prompts to condition the decoder.
We adopt Prefix-Tuning to integrate $P_r$ into each Transformer decoder layer by splitting it into a key prefix $P_K = P_r[:L_p/2,:,:]$ and a value prefix $P_V = P_r[L_p/2:,:,:]$. The prefix tokens $P_K$ and $P_V$ are prepended to the key and value sequences $K$ and $V$, respectively, which are obtained by linearly projecting the decoder’s object queries.
\begin{equation}
{MHA}(Q, K, P_K, V, P_V) = 
{softmax}\left( \frac{Q [P_K, K]^{T}}{\sqrt{d_k}} \right) [P_V, V]
\end{equation}

\noindent\textbf{Inter-Pool Diversity.}
To promote complementary learning between the shared and private pools, we employ the directional decoupled loss~\cite{li2023prompt,zhang2024not}. 
This loss maximizes the angular separation between vectors across the two pools, penalizing pairs with angles below a preset threshold $\theta_{{ddl}}$:
\begin{align}
\theta_{i,j} &= \arccos \left( 
\frac{P_{s,i} \cdot P_{p,j}}
{\| P_{s,i} \| \, \| P_{p,j} \|}
\right) \\
\mathcal{L}_{{DDL}} &= 
\lambda_{{ddl}} \cdot \frac{2}{|N_s||N_p|}
\sum_{i=1}^{|N_s|} \sum_{j=1}^{|N_p|}
\max(0, \theta_{{ddl}} - \theta_{i,j})
\end{align}
where $\theta_{{ddl}} = 90^\circ$ and $\lambda_{{ddl}} = 0.15$. 
This constraint enforces directional separation between the two pools, ensuring effective prompt decoupling.

\subsection{Prototypical Pseudo-Label Generation}
During incremental training at step $t$, generating pseudo-labels with the teacher model $\Phi_{t-1}$ helps maintain supervision consistency across tasks. However, category-wise distribution discrepancies cause confidence-threshold-based methods to struggle in producing reliable pseudo-labels. To address this issue, we propose a Prototypical Pseudo-Label Generation (PPG) module (see Fig.~\ref{fig:ppr}). PPG computes the similarity between class prototypes and candidate pseudo-labels in the embedding space, generating high-quality pseudo-labels for both easy (high-confidence) and hard (low-confidence) samples. This module can be seamlessly integrated into our framework in a plug-and-play manner without any additional post-processing.
\begin{table*}[t]
\centering
\caption{Comparison with state-of-the-art methods on MS-COCO under the multi-step IOD setting, the second-best results are \underline{underlined}.}
\resizebox{1\linewidth}{!}{
\begin{tabular}{c|c|ccc|ccc|ccc}
\toprule
\multirow{2}{*}{{Method}} 
& \multicolumn{1}{c|}{{Task 1}} 
& \multicolumn{3}{c|}{{Task 2}} 
& \multicolumn{3}{c|}{{Task 3}} 
& \multicolumn{3}{c}{{Task 4}} \\ 
\cmidrule(lr){2-11}
& \emph{mAP@C} 
& \emph{mAP@P} & \emph{mAP@C} & \emph{mAP@A}
& \emph{mAP@P} & \emph{mAP@C} & \emph{mAP@A}
& \emph{mAP@P} & \emph{mAP@C} & \emph{mAP@A}\\ 
\midrule
ORE-EBUI~\cite{joseph2021towards}  & 61.4   & 56.5 & 26.1 & 40.6 & 37.8 & 23.7 & 33.7 & 33.6 & 26.3 & 31.8 \\
OW-DETR~\cite{gupta2022ow}   & 71.5   & 62.8 & 27.5 & 43.8 & 45.2 & 24.4 & 38.5 & 38.2 & 28.1 & 33.1 \\
PROB~\cite{zohar2023prob}      & 73.4   & 66.3 & 36.0 & 50.4 & 47.8 & 30.4 & 42.0 & 42.6 & 31.7 & 39.9 \\
CL-DETR~\cite{liu2023continual}   & --     & --   & --   & --   & --   & --   & --   & --   & --   & 39.2 \\
ERD~\cite{feng2022overcoming}       & --     & --   & --   & --   & --   & --   & --   & --   & --   & 35.4 \\
MEPU-FS~\cite{fang2025unsupervised}   & 74.3   & 68.0 & 41.9 & 54.3 & 50.2 & 38.3 & 46.2 & 43.7 & 33.7 & 41.2 \\
SGROD~\cite{he2024recalling}  &73.2   &64.7  &36.7  &50.0   &47.4  &32.4  &42.4  &42.5  &32.6  &40.0 \\
ORTH~\cite{sun2024exploring}      & 71.6   & 64.0 & 39.9 & 51.3 & 52.1 & 42.2 & 48.8 & 48.7 & 38.8 & 46.2\\
MD-DETR~\cite{bhatt2024preventing}   & \underline{78.5}   & 69.1 & \underline{56.5} & \underline{61.2} & \underline{54.6} & \underline{58.3} & \underline{55.4} & \underline{51.5} & \underline{52.7} & \underline{50.2} \\
OWOBJ~\cite{zhang2025open}     & 77.4   & \underline{71.5} & 43.1 & 57.2 & 53.1 & 39.2 & 49.0 & 49.4 & 38.8 & 43.9 \\
\textbf{PDP (Ours)} & \textbf{79.1\rlap{$_{\textcolor{red}{\uparrow 0.6}}$}} 
 & \textbf{77.2\rlap{$_{\textcolor{red}{\uparrow 5.7}}$}} 
 & \textbf{59.9\rlap{$_{\textcolor{red}{\uparrow 3.4}}$}} 
 & \textbf{67.4\rlap{$_{\textcolor{red}{\uparrow 6.2}}$}} 
 & \textbf{65.7\rlap{$_{\textcolor{red}{\uparrow 11.1}}$}} 
 & \textbf{60.5\rlap{$_{\textcolor{red}{\uparrow 2.2}}$}} 
 & \textbf{63.2\rlap{$_{\textcolor{red}{\uparrow 7.8}}$}} 
 & \textbf{61.3\rlap{$_{\textcolor{red}{\uparrow 9.8}}$}} 
 & \textbf{55.8\rlap{$_{\textcolor{red}{\uparrow 3.1}}$}} 
 & \textbf{59.4\rlap{$_{\textcolor{red}{\uparrow 9.2}}$}} \\
\bottomrule
\end{tabular}}
\label{tab:iod_results}
\end{table*}

\noindent\textbf{Class Prototype Space Construction.} 
PPG maintains a prototype for each learned class in the feature space, acting as a stable and generalized feature anchor that captures the semantic core of that class. 
For a new task $t$, prototypes of the newly introduced classes are constructed by extracting object query embeddings $f_i$ of correctly classified instances from the decoder’s final layer. 
These features are stored in a class-specific memory bank $F_c$, and the class prototype $p_c$ is computed as:
\begin{equation}
p_c = \frac{1}{|F_c|} \sum_{f_i \in F_c} f_i
\end{equation}
To ensure prototype stability, both $F_c$ and $p_c$ are updated only during the final epoch of each task, when the feature representations have largely converged.

\noindent\textbf{Hierarchical Validation.}
Building upon these prototypes, PPG performs a hierarchical validation process to refine pseudo-label quality. 
For each image, the teacher model $\Phi_{t-1}$ generates candidate detections associated with confidence scores $s_i$. 
These candidates are then validated in a two-stage manner:  
(1) \textit{Easy Samples:} Detections with high confidence (e.g., $s_i > \tau_h$) are directly accepted as high-precision pseudo-labels.  
(2) \textit{Potential Hard Samples:} For candidates with intermediate confidence ($\tau_l < s_i < \tau_h$), we compute the feature similarity between the object representation and its corresponding class prototype $p_c$. 
If the similarity exceeds a predefined threshold, the detection is retained as a valid hard sample, even if its confidence is relatively low.  
By integrating confidence-based easy samples and prototype-based hard samples, PPG yields a rich and reliable pseudo-label set. 
This dual mechanism effectively balances precision and diversity, thereby maintaining consistent and reliable supervision across tasks. 
Finally, the refined pseudo-labels $Y_{{ppg}}$ are employed to optimize the student model using the MD-DETR~\cite{bhatt2024preventing} detection loss:
\begin{equation}
\mathcal{L}_{{DKD}}(\hat{Y}, Y_{{ppg}}) = \mathcal{L}_{{MD-DETR}}(\hat{Y}, Y_{{ppg}})
\end{equation}

\section{Experiments}

\subsection{Experimental Settings}

\begin{table}[t]
\centering
\small
\setlength{\tabcolsep}{4pt}
\renewcommand{\arraystretch}{1.2}
\caption{Performance comparison under the 40+40 and 70+10 IOD settings, the second-best results are \underline{underlined}.}
\begin{tabular}{c|ccc|ccc}
\hline
\multirow{2}{*}{Method} & \multicolumn{3}{c|}{{40+40}} & \multicolumn{3}{c}{{70+10}} \\
\cline{2-7}
& \textit{AP} & \textit{AP$_{50}$} & \textit{AP$_{75}$} & \textit{AP} & \textit{AP$_{50}$} & \textit{AP$_{75}$} \\

\hline

ABR~\cite{liu2023augmented}  & 34.5 & {57.8} & 35.2 & 31.1 & 52.9 & 32.7 \\
FasterILOD~\cite{peng2020faster}        &20.6 & 40.1 & -- & 21.3 & 39.9 & -- \\
CL-DETR~\cite{liu2023continual} & 42.0 & 60.1 & 45.9 & 40.4 & 58.0 & 43.9 \\
PseudoRM~\cite{yang2023pseudo}  & 25.3 & 44.4 & -- & -- & -- & -- \\
MMA~\cite{cermelli2022modeling}  & 33.0 & 56.6 & 34.6 & 30.2 & 52.1 & 31.5 \\
BPF~\cite{mo2024bridge}  & 34.4 & 54.3 & 37.3 & 36.2 & 56.8 & 38.9 \\
NSGP-RePRE~\cite{wu2025demystifying} & 35.4 & 55.3 & 38.6 & 36.5 & 56.0 & 39.8 \\
PseDet$^{\star}$~\cite{wangpsedet} & \underline{43.5} & \underline{61.5} & \underline{47.2} & \textbf{44.7} & \textbf{62.9} & \textbf{48.6 }\\
\textbf{PDP (Ours)} & \textbf{43.8} & \textbf{62.0} & \textbf{47.5} & \underline{42.9} & \underline{61.1} & \underline{47.1} \\
\hline
\end{tabular}
\label{tab:comparison_40_70}
\end{table}

\begin{table*}[t]
\centering
\small
\caption{Comparison of different methods on the PASCAL VOC dataset under three IOD settings (10+10, 15+5, and 19+1), the second-best results are \underline{underlined}.}
\renewcommand{\arraystretch}{1.1}
\setlength{\tabcolsep}{6pt}
\begin{tabular}{c|ccc|ccc|ccc}
\toprule
\multirow{2}{*}{{Method}} &
\multicolumn{3}{c|}{{10+10}} &
\multicolumn{3}{c|}{{15+5}} &
\multicolumn{3}{c}{{19+1}} \\
\cmidrule(lr){2-4} \cmidrule(lr){5-7} \cmidrule(lr){8-10}
 & \emph{mAP@P} & \emph{mAP@C} & \emph{mAP@A} 
 & \emph{mAP@P} & \emph{mAP@C} & \emph{mAP@A}
 & \emph{mAP@P} & \emph{mAP@C} & \emph{mAP@A} \\
\midrule
ILOD~\cite{shmelkov2017incremental}           & 63.2 & 63.2 & 63.2 & 68.3 & 58.4 & 65.8 & 65.8 & 62.7 & 68.2 \\
Faster ILOD~\cite{peng2020faster}    & 69.8 & 54.5 & 64.2 & 71.6 & 56.9 & 67.9 & 68.9 & 61.1 & 68.5 \\
ORE-EBUI~\cite{joseph2021towards}       & 60.4 & 68.8 & 64.7 & 71.8 & 58.7 & 68.5 & 69.4 & 61.0 & 68.8 \\
OW-DETR~\cite{gupta2022ow}         & 63.5 & 67.9 & 65.7 & 72.2 & 58.9 & 69.4 & 70.2 & 62.0 & 70.4 \\
PROB~\cite{zohar2023prob}            & 66.0 & 67.2 & 66.5 & 73.2 & 60.3 & 70.1 & 73.9 & 48.5 & 72.6 \\
ABR~\cite{liu2023augmented}             & 71.2 & 72.8 & 72.0 & 73.0 & 63.4 & 72.7 & 74.5 & 63.5 & 74.1 \\
BPF~\cite{mo2024bridge}             & 71.7 & 74.0 & 72.9 & 74.0 & 63.2 & 72.7 & 74.8 & 63.5 & 74.1 \\
MD-DETR~\cite{bhatt2024preventing}         & 73.1 & \underline{77.5} & 73.2 & 77.4 & \underline{69.4} & \underline{76.7} & \underline{76.8} & 67.2 & \underline{76.1} \\
RGR~\cite{zhang2025revisiting}             & \underline{75.4} & 76.3 & \underline{75.8} & 75.6 & \underline{69.4} & 73.4 & 75.8 & 67.4 & 75.4 \\
NSGP-RePRE~\cite{wu2025demystifying}      & 75.3 & 72.7 & 74.0 & \underline{77.5} & 61.8 & 73.6 & 76.3 & \underline{69.0} & 76.0 \\
\textbf{PDP (Ours)} & \textbf{81.3\rlap{$_{\textcolor{red}{\uparrow 5.9}}$}} 
 & \textbf{79.4\rlap{$_{\textcolor{red}{\uparrow 1.9}}$}} 
 & \textbf{78.7\rlap{$_{\textcolor{red}{\uparrow 2.9}}$}} 
 & \textbf{80.5\rlap{$_{\textcolor{red}{\uparrow 3.0}}$}} 
 & \textbf{79.4\rlap{$_{\textcolor{red}{\uparrow 10.0}}$}} 
 & \textbf{78.0\rlap{$_{\textcolor{red}{\uparrow 1.3}}$}} 
 & \textbf{79.7\rlap{$_{\textcolor{red}{\uparrow 2.9}}$}} 
 & \textbf{70.1\rlap{$_{\textcolor{red}{\uparrow 1.1}}$}} 
 & \textbf{79.4\rlap{$_{\textcolor{red}{\uparrow 3.3}}$}} \\
\bottomrule
\end{tabular}
\label{tab:VOC}
\end{table*}
\noindent\textbf{Datasets and Metric.} 
We evaluate our approach on two widely used benchmarks: {MS-COCO}~\cite{lin2014microsoft} and {PASCAL VOC}~\cite{everingham2010pascal}. Training is conducted on the official training sets, and evaluation is performed on the MS-COCO validation set and the PASCAL VOC test set, following the {OW-DETR}~\cite{gupta2022ow} protocol.

We report {COCO mAP@IoU=0.5} as the primary evaluation metric. To further assess the trade-off between stability and plasticity in IOD, we adopt three complementary metrics from MD-DETR:

\begin{equation}
\begin{aligned}
{mAP@P} &= {mAP}_{{IoU}=0.5}(\mathcal{C}^{T_1}, \ldots, \mathcal{C}^{T_{t-1}}), \\
{mAP@C} &= {mAP}_{{IoU}=0.5}(\mathcal{C}^{T_t}), \\
{mAP@A} &= {mAP}_{{IoU}=0.5}(\mathcal{C}^{T_1}, \ldots, \mathcal{C}^{T_t}).
\end{aligned}
\label{eq:map_metrics}
\end{equation}

${mAP@P}$ measures performance on previous classes, indicating stability (knowledge retention); 
${mAP@C}$ evaluates performance on current classes, reflecting plasticity (adaptation to new knowledge); 
and ${mAP@A}$ provides an overall measure of continual detection ability over all observed classes.

\noindent\textbf{Implementation Details.} 
Our method is built upon Deformable-DETR~\cite{zhu2020deformable} and implemented using the official files provided by the HuggingFace repository. 
In our experiments, we employ 100 shared prompt tokens, while the number of private prompt tokens is set to match the total number of categories in each dataset (e.g., 80 for COCO and 20 for Pascal VOC). 
The loss weighting coefficients $\lambda_{{ddl}}$, $\lambda_{Q}$ are set to 0.15 and 0.1 respectively. 
The confidence thresholds $\theta_h$ and $\theta_l$ are set to 0.5 and 0.2, respectively, and the prototype similarity threshold $\theta_s$ is fixed at 0.5.
\begin{table*}[t]
\centering
\small
\caption{Ablation study of Private Pool, Share Pool, and PPG modules.}
\renewcommand{\arraystretch}{1.1}
\setlength{\tabcolsep}{4pt}
\begin{tabular}{cccc|cccccccccc}
\toprule
\multirow{2}{*}{PP} & \multirow{2}{*}{SP} & \multirow{2}{*}{PPG} & \multirow{2}{*}{$L_{ddl}$}&
\multicolumn{1}{c}{Task1} & \multicolumn{3}{c}{Task2} &
\multicolumn{3}{c}{Task3} & \multicolumn{3}{c}{Task4} \\
\cmidrule(lr){5-5}\cmidrule(lr){6-8} \cmidrule(lr){9-11} \cmidrule(lr){12-14}
& &  && \emph{mAP@C}  &
\emph{mAP@P} & \emph{mAP@C} & \emph{mAP@A} &
\emph{mAP@P} & \emph{mAP@C} & \emph{mAP@A} &
\emph{mAP@P} & \emph{mAP@C} & \emph{mAP@A} \\
\midrule
\cmark &  &  & & 78.6 & 66.4 & 57.1 & 60.8 & 52.7 & 57.3 & 53.5 & 46.0 & 52.5 & 46.0 \\
% \cmark &\cmark  &  & & 78.4 & 75.2 & 58.8 & 65.7 & 52.7 & 57.3 & 53.5 & 46.0 & 52.5 & 46.0 \\
\cmark & \cmark &  &\cmark& 79.0 & 74.9 & 58.9 & 65.6 & 63.5 & 59.8 & 61.6 & 56.9 & 52.5 & 55.1 \\
 \cmark&  & \cmark & &  79.2 & 77.2 & 58.8 & 66.7 & 64.5 & 58.9 & 62.0 & 59.9 & 55.2 & 58.3 \\
\cmark & \cmark & \cmark & &78.9  & 77.2 &59.5 &67.0 &65.2 &60.0 &62.7 &61.0 &55.3 & 59.0 \\
\cmark & \cmark & \cmark &\cmark& 79.1 & 77.2 & 59.9 & 67.4 & 65.7 & 60.5 & 63.2 & 61.3 & 55.8 & 59.4 \\
\bottomrule
\end{tabular}
\label{tab:ablation}
\end{table*}

\begin{table*}[t]
\centering
\small
\caption{The effect of different threshold parameters $\theta_l$, $\theta_h$, and $\theta_s$.}
\renewcommand{\arraystretch}{1.05}
\setlength{\tabcolsep}{2.8pt}
\begin{tabular}{ccc|c|ccc|ccc|ccc}
\toprule
\multirow{2}{*}{$\theta_l$} & 
\multirow{2}{*}{$\theta_h$} & 
\multirow{2}{*}{$\theta_s$} &
\multicolumn{1}{c|}{Task 1} &
\multicolumn{3}{c|}{Task 2} &
\multicolumn{3}{c|}{Task 3} &
\multicolumn{3}{c}{Task 4} \\
\cmidrule(lr){4-4} \cmidrule(lr){5-7} \cmidrule(lr){8-10} \cmidrule(lr){11-13}
 & & & \emph{mAP@C} &
 \emph{mAP@P} & \emph{mAP@C} & \emph{mAP@A} &
 \emph{mAP@P} & \emph{mAP@C} & \emph{mAP@A} &
 \emph{mAP@P} & \emph{mAP@C} & \emph{mAP@A} \\
\midrule
-- & 0.5 & -- & \textbf{79.2} & 76.9 & 57.3 & 65.8 & 63.1 & 58.0 & 60.6 & 57.6 & 53.8 & 56.1 \\
0.2 & 0.5 & 0.5 & 79.1 & \textbf{77.2} & 59.9 & \textbf{67.4} & \textbf{65.7} & \textbf{60.5} & \textbf{63.2} & \textbf{61.3} & 55.8 & 59.4 \\
0.2 & 0.5 & 0.6 & 79.1 & 77.1 & 59.6 & 67.0 & 65.3 & 60.1 & 62.8 & \textbf{61.3} & \textbf{55.9} & \textbf{59.5} \\
0.2 & 0.5 & 0.7 & 79.1 & \textbf{77.2} & \textbf{60.0} & 67.3 & 65.4 & 60.1 & 62.8 & \textbf{61.3} & 55.7 & 59.4 \\
\bottomrule
\end{tabular}
\label{tab:ablation2}
\end{table*}
\subsection{Results and Analyses}

\noindent\textbf{MS-COCO dataset.} 
We compare SOTA methods under a multi-step incremental configuration, following prior studies~\cite{bhatt2024preventing}. As shown in Table~\ref{tab:iod_results}, our proposed method PDP consistently surpasses all competing approaches across the four incremental tasks and three evaluation metrics, achieving an overall $mAP@A$ of $59.4\%$ after completing all tasks. 
From the variation in $mAP@P$, it can be observed that PDP achieves the lowest forgetting rate on old knowledge, demonstrating strong resistance to catastrophic forgetting. Meanwhile, PDP also outperforms other methods in terms of $mAP@C$, indicating that the decoupled paradigm fully exploits the potential of the private prompt pool and enhances the diversity of prompt representations. In addition, the unbiased supervision provided by PPG further prevents prompt degradation and stabilizes the learning process.

We further conduct two-step incremental experiments under the 40+40 and 70+10 configurations, where PDP still exhibits competitive performance. Under the 40+40 setting, our method outperforms the SOTA PseDet, while in the 70+10 setting, PDP performs slightly worse than PseDet. It is worth noting that PseDet is not an end-to-end framework — it performs inference after each training stage and applies an additional k-means clustering step to generate pseudo-labels as supervision for the subsequent stage.

\begin{figure}[t]
  \centering
   \includegraphics[width=0.9\linewidth]{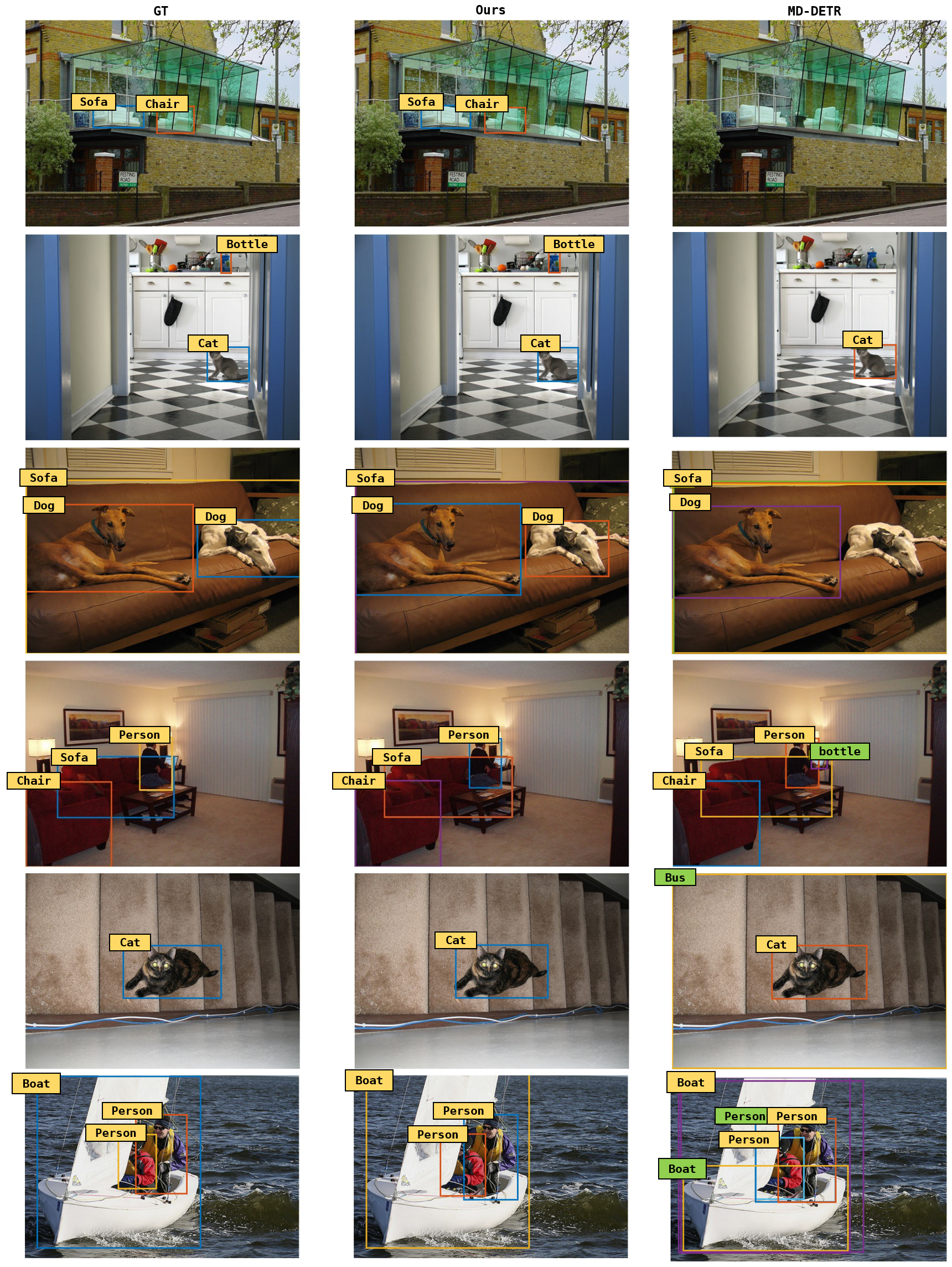}
   \caption{Visualization of old-class detection results on the PASCAL VOC dataset.}
   \label{fig:visual}
\end{figure}

\noindent\textbf{PASCAL VOC dataset.} 
We evaluate our method in three incremental settings: 10+10, 15+5, and 19+1. As shown in Table~\ref{tab:VOC}, PDP consistently outperforms all other methods in terms of $mAP@A$. Compared with the second-best method, PDP achieves an improvement in $mAP@A$ of +2.9\%, +1.3\% and +3.3\% on the 10+10, 15+5 and 19+1 tasks, respectively. We also visualize the performance of PDP in mitigating forgetting, as shown in Fig.~\ref{fig:visual}. Under the 19+1 setting, PDP can accurately detect objects from previously learned classes.

\subsection{Ablation Study}
\noindent\textbf{Analysis of Module components.}
We perform ablation experiments to evaluate the contributions of each component, as shown in Table~\ref{tab:ablation}. The Private Pool (PP) mitigates forgetting of previously learned knowledge by isolating prompts according to task IDs.
The Shared Pool (SP) independently manages task-general prompts, enabling PP to focus on category-specific representations and substantially improving model performance.
This dual-pool paradigm, which explicitly decouples prompt representations, effectively balances model plasticity and stability.
In addition, the proposed PPG improves the retention of previous knowledge by $+13.9\%$ $mAP@P$, significantly enhancing model stability. 
By generating high-quality pseudo-labels, PPG effectively reduces foreground–background conflicts, further improving the plasticity metric $mAP@C$ by $+2.7\%$. 
Finally, the joint use of DDP and PPG achieves the best overall performance, demonstrating their complementary strengths in balancing stability and plasticity.

\noindent\textbf{Analysis of Prototypical Pseudo-label Generation.}
Table~\ref{tab:ablation2} compares our PPG with the fixed-confidence pseudo-labeling method~\cite{liu2023continual} and further analyzes the effect of different similarity thresholds on PPG performance. 
Compared with the fixed-threshold approach, PPG consistently outperforms it across all four tasks, with the most significant gains observed in the final task, achieving improvements of $+3.7\%$ $mAP@P$, $+2.0\%$ $mAP@C$, and $+3.3\%$ $mAP@A$. 
Moreover, the performance of PPG remains stable across three different similarity thresholds, indicating that potentially valuable hard samples maintain high similarity with their corresponding prototypes in the embedding space.

% \specialrule{1pt}{0pt}{0pt}
\begin{figure*}[t]
  \centering
   \includegraphics[width=0.8\linewidth]{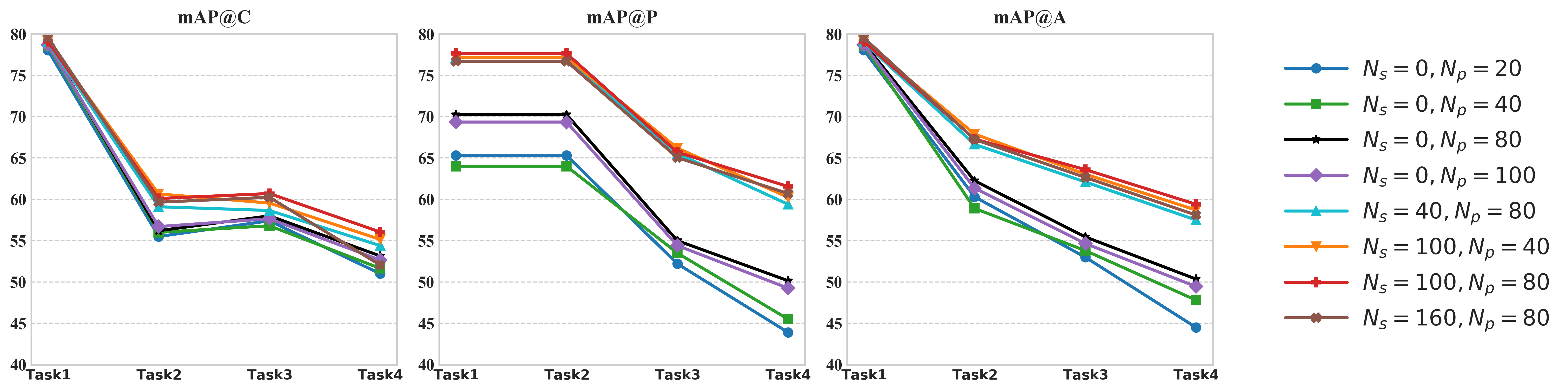}
   \caption{Ablation on shared and private pool sizes. Performance of PDP under different $(N_s, N_p)$ configurations on the COCO incremental detection benchmark, reported in $mAP@C$, $mAP@P$, and $mAP@A$ across sequential tasks.}
   \label{fig:poolsize}
\end{figure*}

\noindent\textbf{Effect of Pool Size.} 
We further conduct an ablation study on the size hyperparameters of the private and shared pools (\(N_p\) and \(N_s\)) to investigate their effects on model performance, as shown in Fig.~\ref{fig:poolsize}. 
The result demonstrates that the combination of \(N_s=100\) and \(N_p=80\) achieves the best overall performance under the COCO multi-step continual detection setting. Specifically, the shared pool size plays a crucial role in maintaining model stability. 
Increasing $N_s$ from 40 to 100 yields notable improvements on subsequent tasks (Task 2-Task 4), demonstrating its importance in capturing and transferring task-general knowledge. 
However, further enlarging $N_s$ to 160 results in a performance drop, indicating that an excessively large shared pool may introduce representational redundancy or optimization difficulties.  In contrast, the private pool size primarily affects model plasticity. 
With $N_s$ fixed at 100, increasing $N_p$ from 40 to 80 consistently enhances performance, particularly on later tasks (e.g., Task~4). This indicates that allocating sufficient parameter space for task-specific prompts enables effective learning of category-discriminative knowledge, enhancing the model’s plasticity.

\section{Conclusion}
\label{sec:conc}
\hspace*{1em}In this paper, we focus on the critical challenge of prompt degradation in IOD. We argue that prompt degradation primarily stems from two root causes: prompt coupling and prompt drift. To address this, we propose PDP, a novel prototype-guided and decoupled prompting framework for IOD. The core of PDP comprises two key innovations. 
First, PDP explicitly decouples task-general and task-specific prompts through shared and private prompt pools. This design enables the shared pool to serve as a stable knowledge foundation for forward transfer, while the private pool focuses on learning discriminative features of new classes, effectively mitigating interference between prompts.
Moreover, PDP designs a prototypical pseudo-label generation module to address prompt drift caused by inconsistent supervision in IOD. By leveraging class prototypes as stable semantic anchors, this module produces reliable pseudo-labels that enhance adaptation to new tasks without compromising prior knowledge.
Extensive experiments demonstrate that PDP achieves state-of-the-art performance across multiple benchmarks, confirming its effectiveness in balancing plasticity and stability.

\noindent\textbf{Acknowledgements.} This work was supported in part by the National Natural Science Foundation of China under Grant 62471394, and U21B2041, 62306241, 62576284.
{
    \small
    \bibliographystyle{ieeenat_fullname}
    \bibliography{main}
}
\clearpage
\setcounter{page}{1}
\maketitlesupplementary

\section{Model Complexity Analysis}
We provide a detailed comparison of model parameters and computational complexity with MD-DETR in Table~\ref{com}. PDP introduces only a marginal increase in parameters (64.9M $\rightarrow$ 66.4M). During inference, the additional computational overhead is negligible (approximately 0.01 GFLOPs). Although a frozen teacher model is employed during training for distillation, it increases GPU memory usage by only approximately 1GB (from 12.3GB to 13.4GB), which does not impose significant practical burden. Importantly, the teacher network is discarded during inference and therefore does not affect deployment efficiency.

\begin{table}[htbp]
\centering
\caption{Model parameters and FLOPs comparison.}
\resizebox{1\linewidth}{!}{
\begin{tabular}{l c c c}
\hline
Method & Params (M) & Training FLOPs (G) & Inference FLOPs (G) \\
\hline
MD-DETR & 64.9 & 166.2 (12.3GB GPU) & 166.17 \\
PDP (Ours) & 66.4 & 332.4 (13.4GB GPU) & 166.18 \\
\hline
\end{tabular}
}
\label{com}
\end{table}
\section{Training Strategy in Incremental Stages}

During each incremental stage (Task $i>1$), distillation guided by PPG is applied to provide supervision for previously learned categories. 
In terms of parameter updates, all parameters in the shared prompt pool are updated throughout training. For the private prompt pool, only the prompts corresponding to the current categories are updated, while the private prompts associated with old and future categories remain frozen. 
This selective update strategy prevents catastrophic forgetting while maintaining adaptability to newly introduced classes.

\section{Additional Results under the $40 + 20 \times 2$ Setting}

To further validate the robustness and generality of PDP, we conduct experiments under the multi-step incremental setting of $40 + 20 \times 2$ on COCO. The results are presented in Table~\ref{422}. 
Under the same Deformable DETR framework, PDP consistently outperforms prior methods, demonstrating strong adaptability in multi-step incremental scenarios.

\begin{table}[htbp]
\centering
\caption{Results under the $40 + 20\times2$ setting on COCO.}
\resizebox{0.8\linewidth}{!}{
\begin{tabular}{l c c c}
\hline
Method & Baseline & $AP$ & $AP_{50}$ \\
\hline
CL-DETR$_{\mathrm{CVPR\,'23}}$ & Deformable-DETR & 35.3 & -- \\
SSDGR$_{\mathrm{CVPR\,'24}}$ & Deformable-DETR & 41.1 & 59.5 \\
DCA$_{\mathrm{AAAI\,'25}}$ & Deformable-DETR & 40.3 & 54.1 \\
PDP (Ours) & Deformable-DETR & \textbf{42.1} & \textbf{60.7} \\
\hline
\end{tabular}}
\label{422}
\end{table}

\section{Upper and Lower Bound Analysis under the $70 + 10$ Setting}

We additionally report results under the $70 + 10$ incremental setting on COCO in Table~\ref{tab:coco_70_10}. For completeness, we provide both a lower bound (direct fine-tuning) and an upper bound (joint training with full access to all data).Direct fine-tuning results in severe forgetting of old categories. In contrast, PDP achieves 43.8\% $AP$ on old categories and approaches the upper bound performance (47.4\% $AP$). In terms of overall performance across all categories, PDP remains only 3.3\% $AP$ below the upper bound, demonstrating effective mitigation of catastrophic forgetting.

\begin{table}[htbp]
\centering
\caption{Performance under the $70 + 10$ setting on COCO.}
\label{tab:coco_70_10}
\resizebox{0.8\linewidth}{!}{
\begin{tabular}{l c c c c}
\hline
\multirow{2}{*}{Method} & \multicolumn{3}{c}{All Categories} & Old Categories \\
 & $AP$ & $AP_{50}$ & $AP_{75}$ & $AP$ \\
\hline
Fine-tune & 4.2 & -- & -- & 0.7 \\
Upper bound & 46.2 & 65.2 & 50.0 & 47.4 \\
PDP (Ours) & \textbf{42.9} & \textbf{61.1} & \textbf{47.1} & \textbf{43.8} \\
\hline
\end{tabular}}
\end{table}


\end{document}